# Are Language Models Models?

Philip Resnik
Department of Linguistics and Institute for Advanced Computer Studies
University of Maryland
resnik@umd.edu

**Abstract**

Futrell and Mahowald claim LMs "serve as model systems", but an assessment at each of Marr's three levels suggests the claim is clearly not true at the implementation level, poorly motivated at the algorithmic-representational level, and problematic at the computational theory level. LMs are good candidates as tools; calling them cognitive models overstates the case and unnecessarily feeds LLM hype.

**For a previous talk closely related to the discussion in this commentary, see: Resnik, P. What Are Large Language Models Models Of?  Presentation at Dagstuhl Seminar 25301, Dagstuhl, Germany. July 21, 2025. https://youtu.be/Gt7IAT9lgrQ. [Video, 26min]**

It's impossible to resist pointing out that in Stanley Kubrick's 1964 film *Dr. Strangelove*, which inspired Futrell and Mahowald's article title, the central theme involved people not thinking things through carefully enough, leading to a global apocalypse. I'm not worried about quite *that* level of catastrophe, mind you, but I do think the article risks further feeding into to a pervasive and ultimately harmful misperception about the relationship between language models (LMs) and human language processing—namely, the too-readily overextended idea of LMs as "model systems" alluded to in the article's abstract.

What do we mean when we say "model system"?  A model, at any level of explanation, is an embodiment of theory that, in addition to capturing generalizations and licensing verifiable inferences and predictions—requirements of any theory—also contains structured correspondences between entities and relationships in the model and corresponding entities and relationships in the real-world system being modeled (Frigg and Nguyen 2017). Notwithstanding its imperfections, Marr (1982) remains a widely accepted scaffolding (Peebles and Cooper, 2015), so let's consider the status of and prospects for LMs as models of language processing at each of Marr's three levels.

First, the **implementation level,** on which we needn't spend much time. The authors argue for a hypothesis-generation role, but they're also clearly aware that functional parallels are not implementation-level models.  Indeed, the connectionism underlying today's LMs was founded on the *abandonment* of implementation-level biological correspondences in favor of function optimization. In Fridman (2021), Jay McClelland recalls Geoff Hinton successfully arguing, "The problem with the way you guys have been approaching this is that you've been looking for inspiration from biology…That's the wrong way to go about it." Consequently, the dominant LM research today demonstrates little substantive engagement with the numerous relevant advances made in implementation-level understanding of the brain since connectionism's loose 20[th]-century inspirations (e.g., Hickok and Poeppel, 2007; Barrett and Satpute, 2013; Hawkins and Ahmad, 2016; Wright et al., 2025; for higher-level reviews see Pulvermüller, 2021, 2023 and Durstewitz et al., 2025).

**Algorithmic/representational level.** The authors' argument here hinges on convergence, that systems solving the same hard problem are driven toward similar mechanisms (Cao and Yamins, 2024). But unlike visual object recognition, a key part of what makes "language" hard is that it decomposes into very distinct subproblems, not merely competing constraints within a single problem. In addition, mechanistic exceptions to convergence are easily found in biology (e.g. sharks and dolphins facing ecologically similar prey-capture problems evolved markedly different mechanisms, relying on electroreception and echolocation, respectively) and in computation (e.g. iterative and recursive computations of the Fibonacci sequence). In medical research, model-organism choices are supported by existing, solidly established mechanistic evidence; e.g., mice and ferrets are preferred in work on tumerogenesis and influenza, respectively—not vice versa—because mice share oncogene pathways with humans and ferrets share lung physiology (Bedell et al., 1997; Belser et al., 2011).

Consider the following scenario. Two aliens from another galaxy, whom I'll dub Abbot and Costello after Villeneuve (2016), show up on earth with their own billions of years of evolution behind them. In a year they absorb nearly everything ever written, and become phenomenal conversationalists; moreover, they generously offer to let human scientists experiment and probe their brain-boxes to our hearts' content. Here's the question: why in the world would one *expect* to find mechanistic correspondences between what's happening in Abbot and Costello's boxes and what's happening in the human cognitive apparatus? Nonetheless, for LMs, the "argument from amazingness" is pervasive—if they are doing something so very human so very well, surely they must be operating under the hood in human-like ways?

As I've highlighted, the argument is unreliable at best, and, moreover, it just isn't necessary in order for LMs to contribute value to computational cognitive modeling. Giulianelli et al. (2024), for example, use LLMs to generate a sample of continuations, in order to operationalize a generalization of surprisal that captures distinctions between anticipatory and responsive online processes in comprehension. This use depends only on generation of plausible text, LMs' undisputed strength (for better or worse; see Resnik, 2025), without requiring claims about their cognitive plausibility. Similarly, LMs can provide useful proxies (with caveats; see Oh and Schuler, 2023) for human comprehenders' next-word probabilities when operationalizing models where such probabilities play a role, e.g. Hahn et al.'s (2022) integration of expectation- and memory-based theories of processing difficulty. LMs are a promising source of world knowledge for hierarchical predictive coding models, which are neurobiologically plausible and have recently been applied to language comprehension (Nour Eddine et al., 2024; Ohams et al., 2026).

**Computational theory level.** Ironically, many who "love the language models" seem to exhibit the same physics envy that you find in Chomskian theory, idealizing a uniform, elegant computation-level paradigm for a wide range of phenomena. But evolution is a scruffy tinkerer, not a neat software engineer (Jacob, 1977). It doesn't re-design from the bottom up when it produces new functionality. For LMs, just as for Chomsky, the drive to explain "du visible compliqué par de l'invisible simple" (Perrin, 1913) creates a problematic gap: even if these architectures can approximate diverse linguistic phenomena, and even if correlations are found between one black box and the other, ultimately there remains the burden of mapping any such computational theory to a biologically plausible algorithmic account of real-time processes distributed over diverse, specialized components. And there are plenty of other less problematic computational-theory games in town (e.g., Eliasmith, 2013; Suárez et al., 2024; Sprevak and Smith, 2023).

At all levels we do need a pluralistic approach, and LMs can be tremendously productive as *tools* in support of developing explanatory models. But LMs themselves are poor candidates as "model systems" at any of Marr's three levels. The distinction matters, for how we prioritize research questions, interpret results, and communicate findings to neighboring fields. For a whole variety of reasons, large language models are the subject of enormous hype. Let's not undermine some of the more nuanced arguments in this paper by overstating the case.

**Acknowledgments**

I'm grateful for helpful comments from Katherine Howitt, Colin Phillips, and Samer Nour Eddine. Any errors or deficiencies are entirely my own.